\documentclass[runningheads,a4paper]{llncs}
\usepackage{amsmath}
\usepackage{amssymb}
\usepackage{graphicx}
\usepackage{array}
\usepackage{booktabs}
\usepackage{url}
\urldef{\mailsa}\path|{alfred.hofmann, ursula.barth, ingrid.haas, frank.holzwarth,|
\urldef{\mailsb}\path|anna.kramer, leonie.kunz, christine.reiss, nicole.sator,|
\urldef{\mailsc}\path|erika.siebert-cole, peter.strasser, lncs}@springer.com|
\newcommand{\keywords}[1]{\par\addvspace\baselineskip
\noindent\keywordname\enspace\ignorespaces#1}

\begin{document}

\mainmatter
\title{Graph-FCN for image semantic segmentation}


\author{ Yi Lu$^1$ \and Yaran Chen $^1$   \and  Dongbin Zhao$^1$  \and Jianxin Chen $^2$
\thanks{This work is supported partly by National Key Research and Development Plan under Grant No.2017YFC1700106, and No. GJHZ1849 International Partnership Program of Chinese Academy of Sciences.}
\authorrunning{Yi Lu, Yaran Chen, Dongbin Zhao, and Jianxin Chen.}
\institute{$^1$State Key Laboratory of Management and Control for Complex Systems\\
Institute of Automation, Chinese Academy of Sciences, Beijing 100190, China\\
University of Chinese Academy of Sciences, Beijing 101408, China\\
$^2$Beijing University of Chinese Medicine, Beijing 100029, China\\
$\textbf{\{}$luyi2017, chenyaran2013,  dongbin.zhao$\textbf{\}}$@ia.ac.cn \  cjx@bucm.edu.cn
}}

\maketitle

\begin{abstract}
Semantic segmentation with deep learning has achieved great progress in classifying the pixels in the image. However, the local location information is usually ignored in the high-level feature extraction by the deep learning, which is important for image semantic segmentation. To avoid this problem, we propose a graph model initialized by a fully convolutional network (FCN) named Graph-FCN for image semantic segmentation. Firstly, the image grid data is extended to graph structure data by a convolutional network, which transforms the semantic segmentation problem into a graph node classification problem. Then we apply graph convolutional network to solve this graph node classification problem. As far as we know, it is the first time that we apply the graph convolutional network in image semantic segmentation. Our method achieves competitive performance in mean intersection over union (mIOU) on the VOC dataset(about 1.34\% improvement), compared to the original FCN model.

\keywords{Graph neural network, Graph convolutional network, Semantic segmentation}
\end{abstract}

\section{Introduction}

The semantic segmentation is an essential issue in the computer vision field, which is much more complex than the classification and detection task\cite{huang_review_2014}. This is a dense prediction task which needs to predict the category of each pixel, namely it needs to learn the object outline, object position and object category from the high-level semantic information and local location information\cite{long2015fully}.

Deep learning-based semantic  segmentation methods, particularly, the convolution neural networks have taken a series of significant progress to this domain. The powerful generalization ability of obtaining the high-level features brings the outstanding performance of the image classification and detection task\cite{Yaran2016wcica,Yaran2016deep}. But the generalization accompanies the loss of local location information, which increases difficulties for dense prediction tasks. The high-level semantic information with a large receptive field corresponds to a small feature map in the convolution neural networks,  which brings the loss of local location information at the pixel-level\cite{Chen2017Multi,Chen2017Multii}.
Many deep learning-based methods have made improvements on this problem, such as full convolution network (FCN)\cite{long2015fully}, Segent\cite{badrinarayanan2017segnet}, Deeplab methods\cite{chen2015semantic,chen2017rethinking,chen2018encoder}. These works use the full connected  layer, dilated convolution, and pyramid structure to lessen the location information loss in extracting high-level features.

In order to solve this problem, firstly, we establish a graph node model for the image semantic segmentation problem .
The graph model methods have been widely used in segmentation problems\cite{gori2005new}. The methods regard the pixels as the nodes, and the dissimilarity between the nodes as the edges. The best segmentation is equivalent to the maximum cut in the graph. And combining the probability and graph theory, the probabilistic graphical model methods, such as Markov random field and conditional random field, are applied to refine the semantic segmentation result\cite{zheng2015conditional,krhenbhl2011efficient}.  These methods model the detected object as the nodes of a graph in the image, and by extracting the relation between the objects to improve the detection accuracy\cite{liu2018structure}.
Compared with the grid structure representation of input data in the deep convolution model, the graph model has a more flexible skip connection, so it can explore a variety of relationships among the nodes in the graph\cite{wang2018local,scarselli2009graph,gilmer2017neural}.

Limited by the amount of calculation, we initialize the graph model by the FCN. The graph model is established on a small size of the image with the nodes annotation initialized by FCN \cite{long2015fully} and the weights of the edges initialized by the Gauss kernel function.

Then we use the graph convolutional network (GCN) to solve this graph model. GCN is one of the state-of-the-art method to deal with graph structure data\cite{kipf2017semi,Defferrard2016Convolutional,li2018deeper}. The node-based GCN uses the message propagation to exchange information between neighbor nodes. This process can extracts the features in a large neighborhood of the graph acted the similar role of convolution and pooling layer in the convolutional network. Because there is no nodes disappear in this process, the node-based GCN expands the receptive field and avoids the loss of local location information.

In this paper, a novel model Graph-FCN is proposed to solve the semantic segmentation problem.
 We model a graph by the deep convolutional network, and firstly apply the GCN method to solve the image semantic segmentation task. The Graph-FCN can enlarge the receptive field and avoid the loss of local location information. In experiments, the Graph-FCN shows outstanding performance improvement compared to FCN.

\section{Problem Formulation}
Semantic segmentation is a pixels classification problem in the image.
In 2015, Jonathan Long et al. used the convolution layer instead of the fully connected layer to establish the end-to-end FCN for pixels classification. The FCN adopts the convolutional layer to extract the local feature on the receptive field. Then it uses the upsampling to restore the feature map to the original size of the image. The model implements pixels-to-pixels mapping, and all the pixels in a single image are propagated forward and backward in parallel. The label image can be obtained by arranging the categories of pixels by pixel position. The input of the FCN is the image $\mathbf{X},\mathbf{X} \in \mathbf{R}^{3 \times w\times h}$, the $w$ is the weight of the image and the $h$ is the height of the image. The output is the predict label image $\mathbf{Y}, \mathbf{Y}\in \mathbf{R}^{w \times h},y_{i,j}\in \mathbf{R}^{w \times h} $. For semantic segmentation task, it is usually uses the cross-entropy loss function of all the pixels in the label image:
\begin{equation}
L_{FCN}=\sum_{i=1}^{w} \sum_{j=1}^{h}-p(y_{i,j}^*)\log{(p(y_{i,j}))}
\end{equation}
where the truth label of the pixel $(i,j)$ of the label image is denoted as $y^*_{i,j}$, and $p(y_{i,j}^*)$ represents the probability of the $y^*_{i,j}$. FCN model can be trained end-to-end by minimizing the cross-entropy loss $L_{FCN}$ using the SGD algorithm.

For the deep learning methods, generalization facilitates identification of deformed objects in image classification and recognition tasks. The pool layer increases the receptive field and decreases the resolution which leads to the loss of pixel position information \cite{long2015fully}.
\begin{figure}
\centering
\includegraphics[height=4cm]{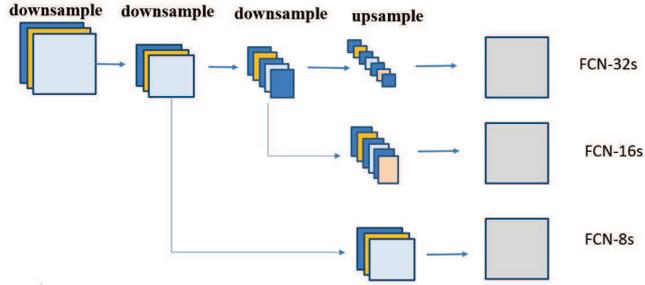}
\caption{The structure of FCN.}
\label{fig:5}
\end{figure}

  FCN introduces the skip connection to fuse feature layers of different scales, as shown in Fig. \ref{fig:5}. %
  Considering that the FCN-16s is just under FCN-8s 0.3 \%  mean intersection over union(mIOU) and has a more concise structure than the FCN-8s, we adopt the FCN-16s as the basic model to initialize the node annotation for the GCN nodes. More details of the nodes describes in section \ref{p:3}.

\section{Graph Model in Semantic Segmentation}

The GCN is designed for solving the learning problem defined on the graph structure data set. The graph structure data can respect as a triple tuple $\mathbf{G}(N, E, U)$. $N$ respects the nodes set of the graph, it is a $|N|*S$ matrix, $|N|$ is the number of the graph nodes, $S$ is the dimension of the node annotation vector. $E$ is the graph edges set. $U$ respects the graph feature, and we omit the $U$ for it not involved in our task. Different from the data representation in Euclidean spatial, the matrix $N$ and edges $E$ are not unique in representation. Matrix $N$ corresponded to $E$, and they are according to the sequence of the nodes. We train the model by supervised learning. The node $n_{j}$ means the node set in graph $j$, $t_{j}$ is the label set to node set $n_{j}$. So the graph model in our task can be shown as the equation (\ref{eg:1}).
\begin{equation}
\begin{aligned}
    &\min_w \ Loss(F_w(\mathbf{G}(N,E)), t)\\
    &s.t.\  \mathbf{G}_j(N,E)\longrightarrow t_j, j\  \epsilon\  T_r.
\end{aligned}
\label{eg:1}
\end{equation}
We use the cross entropy function as the loss function in our model. $T_r$ means the training set.

\subsection{Node}\label{p:3}

\begin{figure}
\centering
\includegraphics[height=3.5cm]{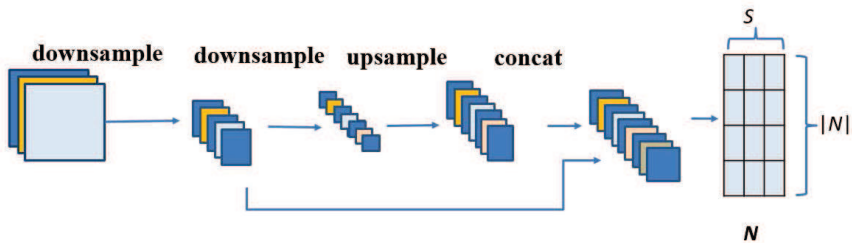}
\caption{The node annotation initialization process. The node annotation is the concatenation of two layers of the FCN-16s.}
\label{fig:1}
\end{figure}

In our model, the node annotations are initialized by the FCN-16s. By the end-to-end training, FCN-16s can get the feature map with a stride of 16 and 32, as shown in Fig. \ref{fig:1}. The feature map with strides 32 can obtain the same size of the feature map with strides 16 by upsampling with the factor 2. The annotation $x_j$ (to node $j$) is initialized by the concatenation of the two feature vectors and the location of each node in the feature map. This annotation contains the extracted features on the local receptive field.
In the training process, we obtain the label of the node by pooling the raw label image.

\subsection{Edge}\label{p:1}

In the graph model, the edge is respected by the adjacent matrix. We assume that each node connects to its nearest $l$ nodes. The connection means that the nodes annotation can be transferred by the edges in the the graph neural network. We give an instance to describe the receptive field in the graph neural network, as shown in Fig.\ref{fig:1}. For example, $l$ is 4.  In the view of the influence of distance on correlation, we adopt the weight adjacent matrix $A$ by the Gauss kernel function.

\begin{figure}
\centering
\includegraphics[height=2.5cm]{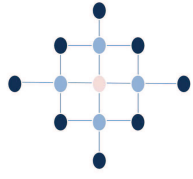}
\caption{The receptive field of a 2-layer GCN when $l$ is 4, which is different from the convolutional layer.}
\label{fig:2}
\end{figure}

\subsection{Training with Graph-FCN }\label{aa}
We use GCN to classify the nodes of the graph model that we have established.
The GCN is one of the deep learning methods to process graph structure\cite{kipf2017semi,Defferrard2016Convolutional}. For a graph the normalized Laplacian matrix $L$ has the form in equation (\ref{eq:2}).
\begin{equation}
L=I - D^{-1/2}AD^{-1/2},
\label{eq:2}
\end{equation}
where matrix $D$ is the diagonal degree matrix, $D_{ii} = \sum_j A_{ij}$ . For the Laplacian matrix $L$ has the orthogonal decomposition
$L = U\Lambda U^T$, the matrix $U$ is orthogonal eigenvectors, the matrix $\Lambda$ is the diagonal matrix of eigenvalues. The graph fourier transform $g_\theta$ is defined as
\begin{equation}
g_\theta(L) *x= g_\theta (U\Lambda U^T)x = U g_\theta(\Lambda)U^T= U diag(\theta)U^Tx.
\label{eq:3}
\end{equation}
Use the Chebyshev polynomials as an approximation of $g_\theta$, we get
\begin{equation}
g_\theta *x \approx \theta_0 x - \theta_1 D^{-1/2}AD^{-1/2} x.
\label{eq:4}
\end{equation}
Due to that $\theta_0 = -\theta_1$ hold in the first order Chebyshev polynomials, we get the equation (\ref{eq:5}),
\begin{equation}
g_\theta *x = \theta_0(I + D^{-1/2}AD^{-1/2})x.
\label{eq:5}
\end{equation}
In order to ensure convergence, the equation Eq.(\ref{eq:6}) exactly is one layer operator in graph convolutional network. This operator takes the role of convolutional and pool layer in the convolutional newtwork and the features are propagated between nodes in this process.
\begin{equation}
X^{k+1} = \hat{A}X^k \Theta ,\hat A =  \hat{D}^{-1/2} (I + A) \hat{D}^{-1/2}),
\label{eq:6}
\end{equation}
where $\hat{D}$ is the degree matrix of $I+A$.

The GCN is a form of Laplacian smoothing. When the messages propagate among the neighbor nodes, the neighbor nodes will tend to have similar features\cite{li2018deeper}. This means that the GCN can not be very deep for the over-smoothing, so we adopt a 2-layers GCN network. The maximum range of node message the current node received can be regarded as the receptive field in the graph. For the instance described in section \ref{p:1}, the size of receptive field is $5\times 32\times 32$, which is five times than that of FCN-16s.  Moreover, there is no nodes disappeared in this progress which means that there is no loss of local location information.

\begin{figure}
\centering
\includegraphics[height=7cm]{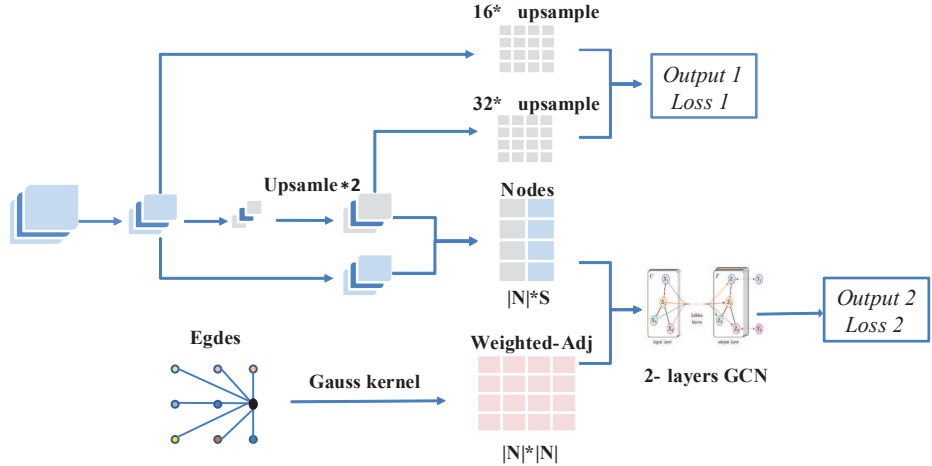}
\caption{The structure of the Graph-FCN. There are two outputs of the model, and and two losses L1 and L2. They share the weights of the feature extracted by convolutional layer.
L1 is calculated by output1 and L2 calculated by the output2. Through minimizing L1 and L2, the FCN-16s can improve performance}
\label{fig:3}
\end{figure}

In the Graph-FCN, the FCN-16s realize the nodes classification and initialization of  the graph model in a small feature map. Meanwhile, the 2-layers GCN gets the classification of the nodes in the graph. We calculate the cross-entropy loss to the both outputs of these two parts. The same as the FCN-16s model, the Graph-FCN is also end-to-end training. The network structure shows in Fig. \ref{fig:3}.
\section{Experiments}
In the experiments, we test our model on the VOC2012 dataset and get the performance improvement than the original FCN model.
\subsection{Implementation}
We take a $366\times 500$ image in VOC dataset as an instance to describe the input and output in detail. In the FCN-16s, after the several pool layers we obtain 512 channels feature map $f1$ and 4096 channels feature map $f2$ of the image. By upsampling, the feature map $f_2$ achieves the same size as feature map $f_1$ ( $4096\times 23\times 32$ ). As described in section \ref{p:3}, we get the the nodes annotation with the size of 4096 + 512 + 2.

In experiments, the input images are the raw images of the VOC data with different size. In order to adapt to the different sizes of the images, we set the batch size 1. The weights of FCN-16s part are initialized by the pre-trained weights, the results of the FCN-16s are shown in the Table \ref{tab:mAP}. The GCN part is initialized randomly. In the first 8,000 iterations, we only adjust the parameters of the GCN part with the learning rate 0.1. Then set the total learning rate 0.00001 to train the whole model with the weight decay 0.1. In the training, we adopt Adam algorithm as the optimizer.
\subsection{Results}
The GCN part in the Graph-FCN model can be regarded as a special loss function. After the model training, the forward output is still the FCN-16s model's output. In the test, the forward part of the Graph-FCN has the same structure as the FCN-16s. But by adding the GCN parts as an additional loss the model the semantic segmentation mIOU has improved 1.34\%.
 \begin{table}[!h]
\centering
 \caption{Graph-FCN vs. FCN-16s  }
 \begin{tabular}{p{2cm}|p{2cm}<{\centering}|p{2cm}<{\centering}|p{2cm}<{\centering}cp{2cm}<{\centering}}
  \toprule
 method($\%$) &mIOU& ACC&  f.w.IU   \\
 \hline
  \midrule
 FCN-16s &64.57 &90.67&84.19\\
 Graph-FCN &65.91 &91.98 & 85.68\\
  \bottomrule
 \end{tabular}
 \label{tab:mAP}
\end{table}

\begin{figure}
\centering
\includegraphics[width=10cm,height=8cm]{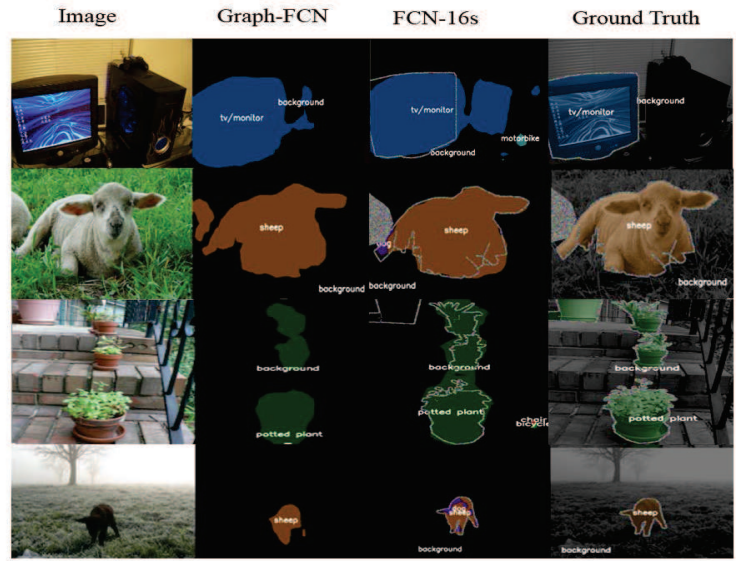}
\caption{The image semantic segmentation results. The second column is the Graph-FCN results. The third column is the FCN-16s results. The fourth column is the ground truth.}
\label{fig:4}
\end{figure}
Fig.4 shows some samples predicted by Graph-Fcn and FCN. From the Fig.4, we can see that the proposed Gra has much smoother results compared with FCN-16. It may be due to that Graph-FCN applies the function of Laplacian smoothing to smooth the predictions. Moreover, the proposed method reduces classification error rate. For example, FCN-16 classfies a part of a sheep as a part of a dog, shown in the second line of Fig.4. It reflects that the Grap-fcn can extract the messages from the neighbour nodes which help the current node classification.

\section{Conclusion}
We model a graph network on the image by the FCN-16s, and propose a Graph-FCN model for semantic segmentation task. The Graph-FCN model can extract the feature on a larger receptive field than the FCN-16s. In the experiment, for the same forward structure, the Graph-FCN achieves a higher mIOU than the FCN-16s, which proves that the Graph-FCN enhance the feature extracting for the pixel classification.

\bibliographystyle{splncs04}
\bibliography{inss}
\end{document}